\documentclass[12pt]{article}
\usepackage[utf8]{inputenc}
\usepackage[T1]{fontenc}
\usepackage{lmodern}
\usepackage{geometry}
\usepackage{graphicx}
\usepackage{tabularx}
\usepackage{booktabs}
\usepackage{array}
\usepackage{hyperref}
\usepackage{longtable}
\usepackage{subcaption}
\usepackage{caption}
\usepackage{amsmath}
\title{
    \LARGE\bfseries
    Intertextual Parallel Detection\\ in Biblical Hebrew:\\
    A Transformer-Based Benchmark\\[1.2ex]
}

\author{David M. Smiley\\University of Notre Dame\\dsmiley@nd.edu}
\date{}

\begin{document}
\maketitle

\begin{abstract}
Identifying parallel passages in biblical Hebrew (BH) is central to biblical scholarship for understanding intertextual relationships. Traditional methods rely on manual comparison, a labor-intensive process prone to human error. This study evaluates the potential of pre-trained transformer-based language models, including E5, AlephBERT, MPNet, and LaBSE, for detecting textual parallels in the Hebrew Bible. Focusing on known parallels between Samuel/Kings and Chronicles, I assessed each model's capability to generate word embeddings distinguishing parallel from non-parallel passages. Using cosine similarity and Wasserstein Distance measures, I found that E5 and AlephBERT show promise; E5 excels in parallel detection, while AlephBERT demonstrates stronger non-parallel differentiation. These findings indicate that pre-trained models can enhance the efficiency and accuracy of detecting intertextual parallels in ancient texts, suggesting broader applications for ancient language studies.
\end{abstract}

\section{Introduction}
\footnote{All code and data can be found in the following Github repository: \url{https://github.com/dmsmiley/detect-bh}}
Identifying parallel passages in biblical Hebrew (BH) texts has long been central to biblical scholarship, serving as a tool for understanding intertextual relationships and theological development \cite{Fewell}. Scholars have traditionally relied on manual comparisons to trace these connections \cite{Harvey, Miller, Schnittjer}. These parallels are not merely textual repetitions; they often involve reinterpretations and recontextualizations, providing insights into the evolution of biblical narratives \cite{Fishbane, Kalimi}.

Traditional methods for detecting these parallels are labor-intensive and susceptible to human error, especially when the connections are implicit, nuanced, or subtle. Even for computational approaches the linguistic complexity of BH, with its rich morphology and intricate verbal structures, exacerbates these challenges \cite{Singh}. These difficulties are magnified when scholars manually analyze texts spanning extensive chronological, thematic, and theological corpora, pushing the limits of attention and leading to inconsistencies in intertextual findings.

Recent advances in natural language processing (NLP) present new possibilities for addressing these limitations. Large language models (LLMs), particularly transformer-based architectures, excel at capturing semantic relationships across vast datasets \cite{Devlin}. These models process large amounts of text with high accuracy, generating rich word embeddings reflecting nuances of meaning and context that traditional methods cannot.

This study evaluates the potential of pre-trained transformer models, such as E5, AlephBERT, MPNet, and LaBSE, to aid in detecting parallel passages in the Hebrew Bible (HB). Focusing on known parallels that Chronicles (Chr) reuses from Samuel (Sam) and Kings (Kgs), I assess each model's ability to generate word embeddings that accurately delineate known parallel and non-parallel verses in the HB. This research not only provides benchmarks of model accuracy for textual similarity tasks, but also lays the groundwork for future NLP applications in biblical studies and broader applications to other ancient literature.

\section{Previous Computational Attempts at Biblical Hebrew Parallels}
Various computational methods have been applied to identifying parallels in BH, though often constrained by the available tools and techniques. Notably, van Peursen and Talstra \cite{vanPeursen} sought to systematically compare parallel texts from 2 Kgs, Isaiah, and Chr using a computer-assisted synopsis. Their approach highlights the difficulty in establishing a rule-based method for identifying parallel texts. Criteria for identifying parallels are often subjective, and proper parameters are opaque.

In recent works, Shmidman, Koppel, and Porat \cite{Shmidman} have significantly refined rule-based approaches for detecting parallels in large Hebrew and Aramaic corpora. Their algorithms are designed to detect approximate matches, which account for rephrasing, orthographic differences, and interpolations. This method has proven effective in identifying textual reuse in complex corpora like the Babylonian Talmud, where near-identical passages often exhibit minor discrepancies.

Shmidman \cite{Shmidman2022} further refined his work by creating a hashing algorithm that breaks each lexeme into the smallest unit for word representation. However, as Shmidman acknowledges, this method is susceptible to numerous false positive matches. The inherent limitation of rule-based methods is their inability to capture deeper semantic relationships, often missing broader contextual nuances in rewritten texts like those in Chr. This underscores the challenges faced by rule-based approaches in understanding the complex semantic landscape of ancient texts.

These frequency- and rule-based approaches rely too heavily on surface-level lexical matching rather than more sophisticated linguistic representations \cite{Mars}. Furthermore, parallel passages are often not exact copies of previous texts, meaning that passages with significant omissions, additions, or rewrites, such as those in Chr, are not accurately captured by these methods. This highlights the need for methods that effectively capture semantics rather than solely relying on superficial matching.

\begin{table}[htbp]
\centering
\begin{tabularx}{\textwidth}{
  @{}>{\raggedright\arraybackslash}p{2cm}
     >{\raggedright\arraybackslash}p{1.8cm}
     >{\raggedright\arraybackslash}X
     >{\raggedright\arraybackslash}p{1.8cm}
     >{\raggedright\arraybackslash}X@{}
}
\toprule
\textbf{Type of Textual Difference} 
 & \textbf{Sam/Kgs Reference} 
 & \textbf{Sam/Kgs Text} 
 & \textbf{Chr Reference} 
 & \textbf{Chr Text} \\
\midrule
Addition and Name Changes 
 & 2 Sam 5:6 
 & The king and his men went to Jerusalem against the Jebusites, the inhabitants of the land.
 & 1 Chr 11:4
 & David and all of Israel went to Jerusalem, which was Jebus. There the Jebusites were the inhabitants of the land. \\[1ex]

Omission 
 & 2 Sam 6:15 
 & David and all of the house of Israel brought up the ark of the covenant\ldots 
 & 1 Chr 15:28 
 & [ ] All of Israel brought up the ark of the covenant\ldots \\[1ex]

Addition and Omission 
 & 2 Kgs 11:4 
 & Jehoiada\ldots took the officers of one hundred from the Carites and the guard, and brought them to him in the house of YHWH, and he cut a covenant with them.
 & 2 Chr 23:1
 & Jehoiada\ldots took the officers of one hundred, Azariah son of Jeroham, Ishmael son of Jehohanan, Azariahu son of Obed, Maasieah son of Adaiahu, and Elishaphat son of Zichri into a covenant agreement with him. \\[1ex]

Omission 
 & 2 Sam 7:14 
 & I will be like a father to him, and he will be like a son to me. Whenever he goes astray, I will punish him with the staff of men and with the beatings of the songs of man.
 & 1 Chr 17:13
 & I will be like a father to him, and he will be like a son to me. [ ] \\
\bottomrule
\end{tabularx}
\caption{Examples of rewriting patterns from Kalimi's work \cite{Kalimi} employed by the author of Chronicles to reshape earlier parallel passages.}
\label{tab:kalimi}
\end{table}

Recent advances, particularly in transformer-based models, offer a solution to challenging parallels, like those shown in Table 1. Unlike frequency-based methods, transformers can generate verse-level embeddings based on a word's relationship to its context, thus capturing semantic variation \cite{Vaswani}. This ability to contextualize words semantically makes them particularly suited to uncovering parallel passages in BH.

Despite their potential, no systematic evaluation has been conducted to benchmark these modern language models on BH texts. While models like AlephBERT \cite{Seker} have been trained on vast amounts of modern Hebrew (MH), their effectiveness on ancient Hebrew remains untested. The linguistic differences between MH and BH, combined with the absence of BH-specific language models, create a significant gap in our understanding of how well these pre-trained models can handle ancient language tasks.

This study addresses this gap by providing the first comprehensive benchmark of transformer-based models for detecting parallels in biblical texts. By leveraging known parallels between Sam/Kgs and Chr, I systematically evaluate how different pre-trained LLMs perform when creating embeddings for BH verses. This benchmark study not only advances NLP techniques for ancient texts, but also provides crucial empirical data for scholars seeking to apply these tools to biblical scholarship without developing models from scratch.

\section{Objectives}
The primary objective of this study is to assess the effectiveness of transformer-based language models in identifying parallel passages within BH texts. Specifically, I will evaluate how well pre-trained models like E5, AlephBERT, MPNet, and LaBSE capture the relationships between parallel texts, particularly those between Sam/Kgs and Chr.

By comparing how well these models create word embeddings distinguishing true parallel passages from non-parallels, I can estimate their suitability for finding unknown connections traditionally identified through manual methods in biblical scholarship. This demonstrates how transformer-based models can enhance the accuracy and efficiency of detecting intertextual parallels in ancient texts. In addition to assessing the strengths and limitations of each model, this study contributes to the broader effort to integrate advanced NLP techniques into the field of digital humanities. The findings offer an empirical method for detecting previously unknown parallel and intertextual connections in historical texts.

\section{Data and Models}
\subsection{Dataset}
The Hebrew text for this study is drawn from the Biblia Hebraica Stuttgartensia Amstelodamensis (BHSA) corpus, as compiled by the Eep Talstra Centre for Bible and Computer at Vrije Universiteit Amsterdam \cite{vanPeursen}. Additionally, 558 verses from Chr have recognized parallels in Sam/Kgs \cite{Enders}, ensuring a consistent, empirical set of passages for evaluating the transformer-based models' performance.

\subsection{Pre-Trained Models}
Four pre-trained transformer models were used, each selected for their strengths in text classification and documentary similarity. Some are optimized for their scale and multilingual capabilities like Multilingual E5 \cite{Wang}. Another has its strength in representing larger sentences and paragraphs such as MPNet \cite{Song}. LaBSE is a language agnostic approach to embeddings \cite{Feng}. In contrast the final model, AlephBERT, is specifically pre-trained on MH \cite{Seker}. Selecting robust, diverse models allows observation of which approaches best generalize embeddings for text similarity in BH.

\section{Model Evaluation}
\subsection{Cosine Similarity}
Cosine similarity is the preferred measure for evaluating vector proximity in text similarity tasks by calculating the cosine of the angle between two vectors in high-dimensional space. This score is particularly useful for analyzing embeddings generated by LLMs, as it captures both syntactic and semantic relationships between words \cite{Gomez}.

In the context of BH, parallel passages often display variations in word choice, spelling, and theological interpretation, as shown above in Table 1 \cite{Kalimi}. Therefore, a later account is not always a simple “copy and paste” version of its earlier parallel text, making frequency-based methods insufficient for finding parallels. By representing each passage as a vector, cosine similarity quantifies the semantics of a text, regardless of superficial additions, omissions, or spelling differences.

For this study, cosine similarity was computed in two ways. First, each passage in Chr was compared to its known parallel in the books of Sam/Kgs, generating a cosine similarity score:
\[
\mathrm{Cosine}(v^{\text{Chr}}_i, v^{\text{Sam/Kgs}}_j)
\]
Second, each passage from Chr was compared to every other verse in Sam/Kgs that is not a known parallel to compute the mean non-parallel cosine similarity:
\[
\mu_{\text{NonParallel}}(v_i^{\text{Chr}}) = \frac{1}{N} \sum_{j=1}^N \text{Cosine}(v_i^{\text{Chr}}, v_j^{\text{Sam/Kgs}}), \quad \text{where } v_j \notin \text{Parallels}(v_i)
\]
This approach evaluates how well the models detect true parallels and their ability to avoid false positives by assigning high similarity to unrelated passages.

\subsection{Parallel and Non-Parallel Verse Cosine Similarities}
The table below summarizes the performance of the pre-trained models, including the mean cosine similarity for parallel and non-parallel passages, and the statistical significance (p-value) of the difference between mean cosine similarities for parallel and non-parallel texts. The Appendix contains the histogram of the parallel cosine similarity scores for each model.

\begin{table}[htbp]
\centering
\label{tab:cosine}
\begin{tabularx}{\textwidth}{
  @{}>{\raggedright\arraybackslash}p{2.8cm}
     >{\raggedright\arraybackslash}p{2.0cm}
     >{\raggedright\arraybackslash}p{2.0cm}
     >{\raggedright\arraybackslash}p{2.7cm}
     >{\raggedright\arraybackslash}p{2.5cm}
     >{\raggedright\arraybackslash}p{2.5cm}@{}
}
\toprule
\textbf{Model} & \textbf{Mean Cosine (Parallel)} 
               & \textbf{Mean Cosine (Non-Parallel)} 
               & \textbf{P-value} 
               & \textbf{\% Cosine $\geq$ 95\%} 
               & \textbf{\% Cosine $\geq$ 98\%} \\
\midrule
E5         & 0.966 & 0.882 & $1.06\times10^{-284}$ & 75.0\% & 40.11\% \\
AlephBERT  & 0.914 & 0.638 & $9.98\times10^{-314}$ & 44.78\% & 17.27\% \\
MPNet      & 0.903 & 0.649 & $1.61\times10^{-231}$ & 46.76\% & 25.72\% \\
LaBSE      & 0.828 & 0.375 & $2.48\times10^{-270}$ & 26.07\% & 11.33\% \\
\bottomrule
\end{tabularx}
\caption{Cosine similarity scores and p-value from a t-test comparing the means of cosine similarity scores.}
\end{table}

E5 consistently outperforms the other models, achieving a high mean cosine similarity of 0.966 for parallel passages, with 75.0\% of passages exceeding a 95\% similarity. However, its high similarity score for non-parallel passages (0.882) suggests an inability to properly delineate between the parallels and non-parallels. AlephBERT, while designed for Hebrew, underperforms relative to E5, with a mean cosine similarity of 0.914 for parallel passages. Only 17.27\% of parallels exceeded 98\% similarity. However, AlephBERT demonstrates stronger separation between parallel and non-parallel passages (mean cosine for non-parallels: 0.638), suggesting that it might be properly fitting better than E5.

MPNet and LaBSE exhibit the weakest performance with lower mean cosine similarities for parallel passages (0.903 and 0.828, respectively) and less distinct separation between parallel and non-parallel passages. This likely stems from their lack of optimization for Hebrew, which affects their ability to capture the linguistic nuances of the text.
A t-test confirms that the observed differences between mean parallel and non-parallel cosine similarities are significant across all models, with p-values below 1e-100. For instance, E5's p-value of 1.06e-284 indicates a highly significant distinction between the mean cosine score for parallel and non-parallel texts. Therefore, the models can consistently achieve similarity scores distinguishing parallel and non-parallel passages, despite not being optimized for BH.

\subsection{Wasserstein Distance Analysis}
To further assess model performance, the Wasserstein Distance, a distribution-based metric, was employed to measure the separation between cosine similarity distributions for parallel and non-parallel passages. A higher Wasserstein Distance indicates greater separation and a stronger proclivity to distinguish between parallel and non-parallel verses \cite{Leo}. 
\begin{table}[htbp]
\centering
\label{tab:wasserstein}
\begin{tabular}{@{}lc@{}}
\toprule
\textbf{Model} & \textbf{Wasserstein Distance} \\ 
\midrule
E5         & 0.0812 \\
AlephBERT  & 0.2764 \\
MPNet      & 0.2540 \\
LaBSE      & 0.4532 \\
\bottomrule
\end{tabular}
\caption{Wasserstein distance scores comparing the output distributions of different models.}
\end{table}

The results reveal important insights about model characteristics. E5 exhibits the lowest Wasserstein Distance (0.0812), indicating significant overlap between the cosine similarity distributions for parallel and non-parallel passages. This confirms E5's tendency to broadly assign high similarity scores for true parallels; however, it does so at the expense of also assigning high cosine similarity scores to false positives, as reflected in the limited distribution separation.

In contrast, AlephBERT shows a larger Wasserstein Distance (0.2764), indicating better separation between parallel and non-parallel passages. While AlephBERT's overall cosine similarity for parallel passages is lower than E5's, its greater separation suggests that it is less prone to false positives.

MPNet and LaBSE also exhibit larger Wasserstein Distances (0.25 and 0.45, respectively), reflecting better distinction between parallel and non-parallel passages. However, their lower overall performance in mean cosine similarity limits their effectiveness for detecting true parallels. MPNet has a slightly worse, but similar performance to AlephBERT on both measures. While LaBSE has the largest Wasserstein Distance, as the next section will confirm, it may be too conservative in identifying parallels, potentially missing subtle intertextual connections.

\subsection{Classification Metrics}
Cosine similarity was also used to find the closest match for each passage, helping to identify which specific text was most similar to our parallel text in Chr. This nearest-neighbor approach evaluates whether the models are finding true parallels or capturing incorrect matches. To further assess model performance, classification reports were created, focusing on precision, recall, and F1-score. The micro-averaged metrics \cite{Rainio} clearly illustrate how well each model distinguished between true parallels and unrelated passages by finding the closest cosine similarity match for each verse in Chr.

\begin{table}[htbp]
\centering
\label{tab:classification}
\begin{tabular}{@{}lccc@{}}
\toprule
\textbf{Model} & \textbf{Recall} & \textbf{Precision} & \textbf{F1-Score} \\ 
\midrule
E5         & 0.85 & 0.92 & 0.88 \\
AlephBERT  & 0.82 & 0.92 & 0.87 \\
MPNet      & 0.68 & 0.90 & 0.78 \\
LaBSE      & 0.72 & 0.86 & 0.78 \\
\bottomrule
\end{tabular}
\caption{Classification report evaluating the accuracy of identifying closest parallel passages based on cosine similarity scores.}
\end{table}

E5 achieved a micro average precision of 0.92 and a recall of 0.85, indicating that while the majority of detected parallels were correct, around 15\% of true parallels were not identified as the closest parallel text based on cosine similarity. This resulted in a strong F1-score of 0.88, demonstrating effective performance overall.

AlephBERT demonstrated similar performance, with a micro average precision of 0.92 and a recall of 0.82, resulting in an F1-score of 0.87. The metrics for AlephBERT are close to those of E5, suggesting that both models perform similarly in identifying true parallels and avoiding false positives for the closest parallel detected.

MPNet and LaBSE showed lower effectiveness overall. They performed well with F1-scores of 0.78, but still fell short of the performance levels seen with E5 and AlephBERT.

These classification metrics complement cosine similarity and Wasserstein Distance analyses by providing insights into the balance between precision and recall for each model. E5 and AlephBERT both effectively detect parallels, with similar precision and recall metrics, indicating comparable performance. MPNet and LaBSE demonstrate limited reliability, suggesting a combined approach using E5 and AlephBERT may yield the best results for identifying intertextual parallels in BH texts.

\subsection{Model Conclusions}
Based on the evaluation of pre-trained models, E5 and AlephBERT stand out as the most effective models for detecting true parallel passages in BH texts. E5 excels at detecting explicit parallels, achieving high mean cosine similarity for known passages. However, its performance also reveals a tendency to catch false positives, with high similarity scores for non-parallel texts, as reflected in its relatively low Wasserstein Distance. This makes E5 highly capable of identifying parallels but a bit less effective in distinguishing true from false positives.

AlephBERT, while slightly behind E5 in mean cosine similarity for parallel passages, shows a slight edge in avoiding false positives. Its lower similarity for non-parallel texts and higher Wasserstein Distance indicate better separation between related and unrelated passages. This makes AlephBERT particularly suitable when minimizing incorrect matches is a priority, providing a complementary strength to E5.

MPNet and LaBSE showed weaker performance in identifying true parallels, largely due to their lower optimization for BH. These models exhibited lower cosine similarities and greater difficulty in distinguishing parallel from non-parallel texts. Therefore, they should be avoided in their current forms for creating word embeddings for BH.

These findings suggest that using both the E5 and AlephBERT models as a check and balance of one another offers a promising solution for enhancing parallel detection accuracy. Future work should focus on fine-tuning these pre-trained models for BH to maximize their complementary strengths, potentially revolutionizing the identification of intertextual parallels, not only in BH, but also in other ancient texts. This approach enables scholars to utilize advanced NLP tools without developing new models from scratch, instead focusing on optimizing and adapting existing models for specific historical and linguistic contexts.

\section{Future Directions}
This study demonstrates the capability of pre-trained transformer-based models to identify parallel verses in BH using document similarity measures such as cosine similarity and Wasserstein Distance. E5 and AlephBERT show significant promise in detecting these passages despite differences in lexical choice or sentence structure. Although not perfect, adapting and fine-tuning these models can bypass the issues of building a BH language model from scratch.

Expanding this research to other ancient languages, such as Syriac, Greek, or Latin, could provide further insights into intertextuality in other areas of ancient language study. If these models generalize well with BH, then in theory they could also work well in other languages on which they are not explicitly trained. Developing embeddings from pre-trained models for ancient texts could transform the study of literary connections in the ancient world, offering new tools for scholars in the digital humanities.

\section{Limitations}
While this study demonstrates the promise of pre-trained transformer-based models for detecting biblical parallels, several limitations should be acknowledged. First, this evaluation focuses exclusively on narrative texts, specifically known parallels between Sam/Kgs and Chr. The HB encompasses numerous genres, like poetry, wisdom literature, prophetic oracles, and legal codes, each with unique ways of displaying intertextual connections. The performance of these models on something like poetic parallelism in the Psalter, the covenant code in Exodus, or prophetic sayings of Jeremiah remains untested, thus limiting the generalizability of our findings across the full spectrum of biblical literature.

Second, the embeddings generated by these pre-trained models, while effective for similarity detection, remain largely opaque and uninterpretable. This “black box” nature of all language models built on the transformer architecture prevents scholars from understanding why certain passages are deemed similar, potentially obscuring important linguistic or theological nuances that traditional philological methods would capture. This limitation is particularly significant for biblical studies, where understanding the specific mechanisms of textual reuse is often as important as identifying the parallels themselves.

Finally, this study evaluates existing pre-trained models rather than developing a BH-specific language model. While our approach demonstrates that adaptation of existing models is viable and practical, a purpose-built model trained on BH corpora might better capture the unique morphological, syntactic, and semantic features of ancient Hebrew texts, potentially improving both accuracy and interpretability. Unfortunately, building a language model from scratch is likely infeasible due to the HB's small corpus and the data needs of transformers; however, fine-tuning models like AlephBERT or E5 for textual similarity on BH remains a productive avenue for future research.

\clearpage
\appendix
\section*{Appendix}

Each graph contains the distribution of cosine similarity scores for all known parallels (in blue) alongside the distribution of non-parallel cosine similarity scores (in red) according to the embeddings created by the different models.

\begin{figure}[htbp]
\centering

\begin{subfigure}{0.45\textwidth}
    \centering
    \includegraphics[width=\linewidth]{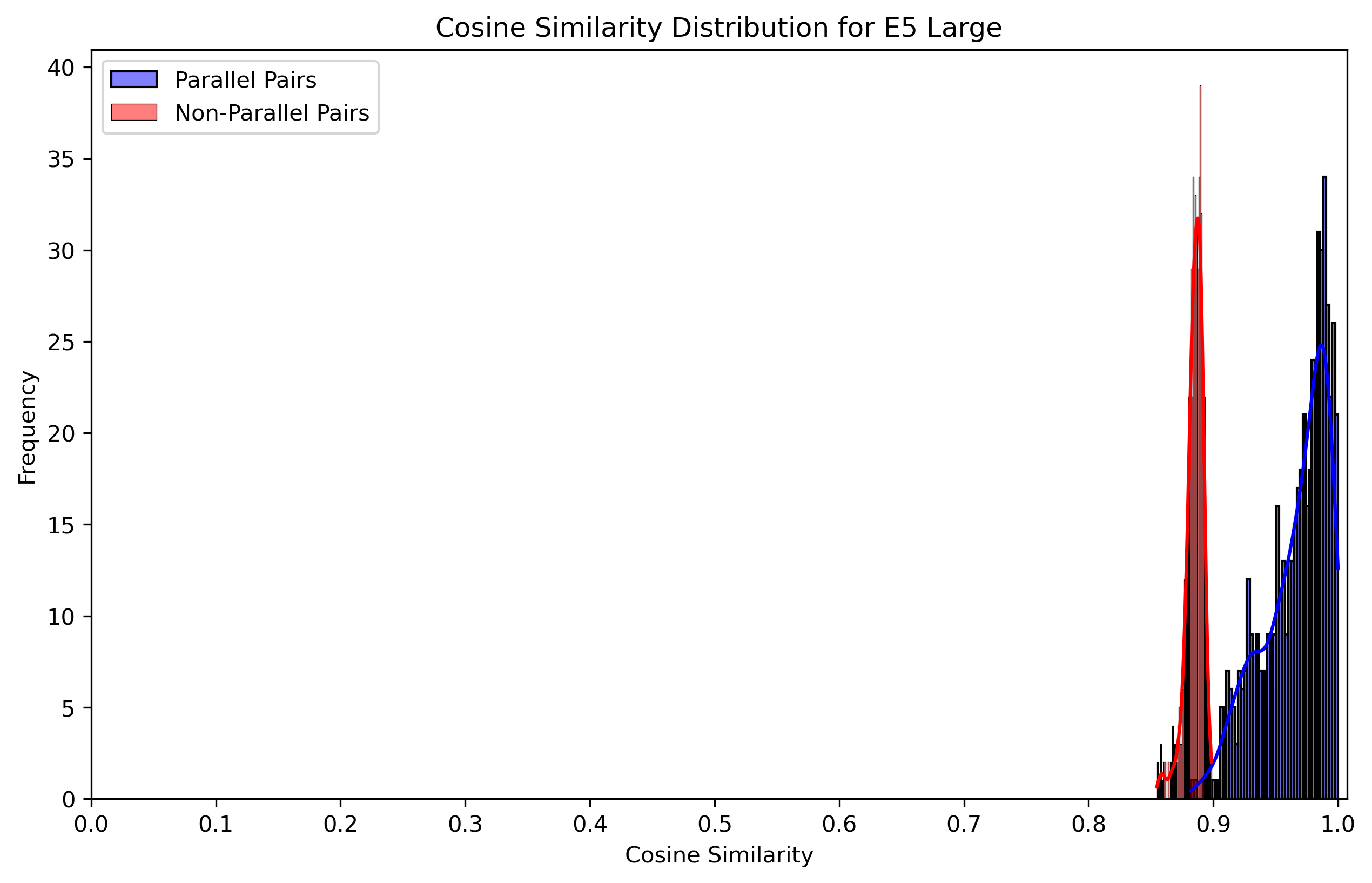}
    \caption{E5}
    \label{fig:e5_dist}
\end{subfigure}
\hfill
\begin{subfigure}{0.45\textwidth}
    \centering
    \includegraphics[width=\linewidth]{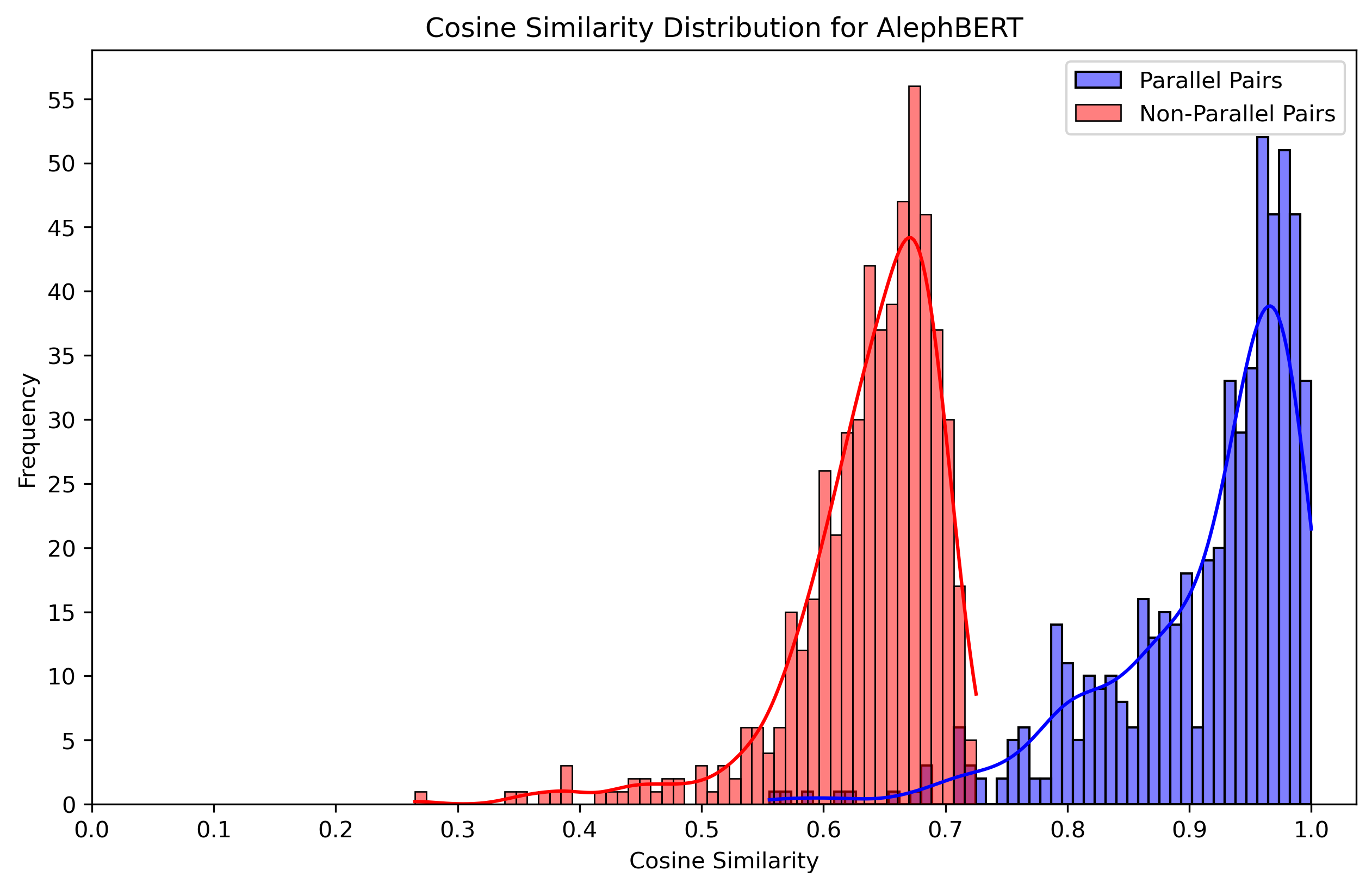}
    \caption{AlephBERT}
    \label{fig:alephbert_dist}
\end{subfigure}

\vspace{1em}

\begin{subfigure}{0.45\textwidth}
    \centering
    \includegraphics[width=\linewidth]{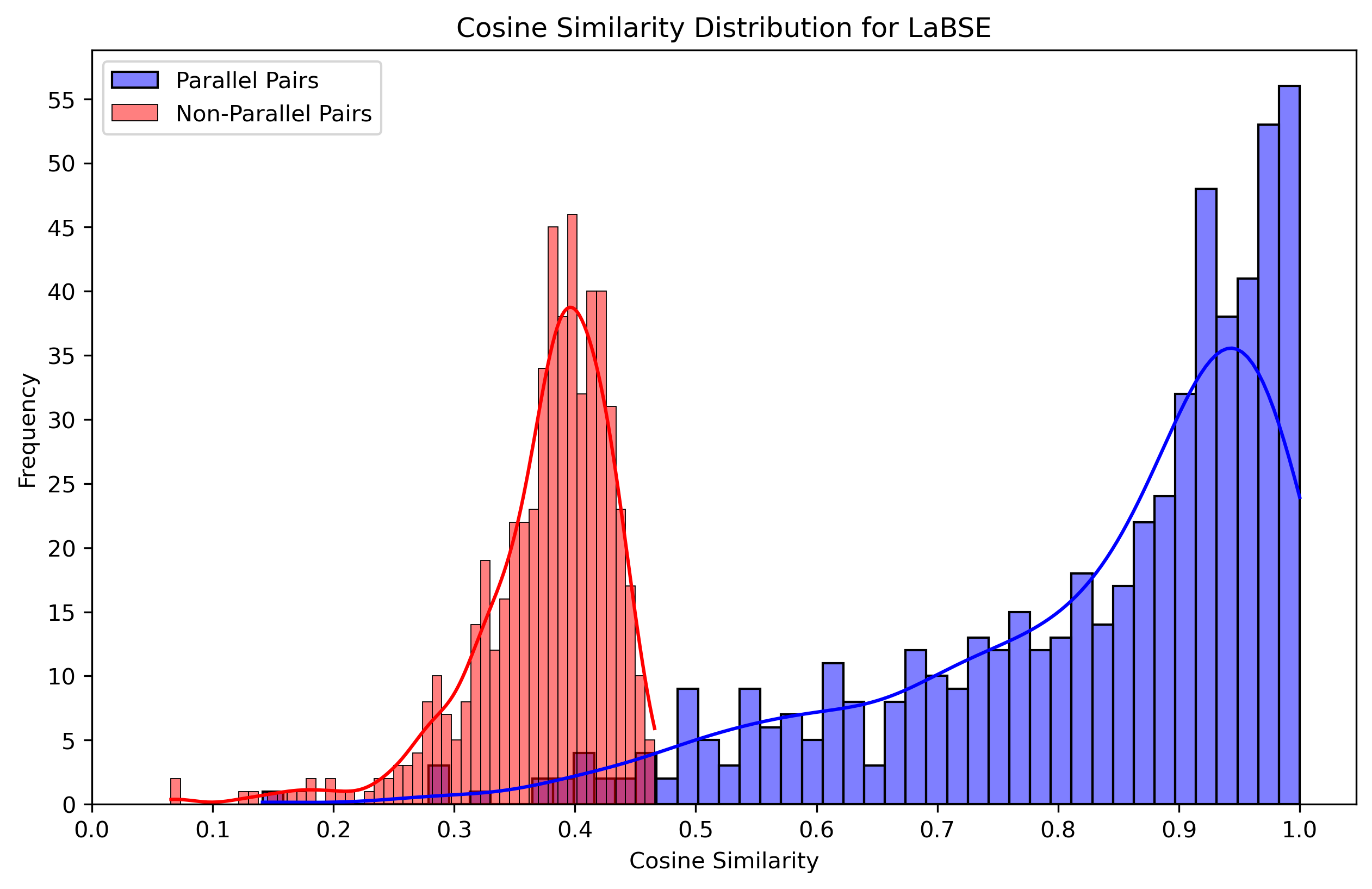}
    \caption{LaBSE}
    \label{fig:labse_dist}
\end{subfigure}
\hfill
\begin{subfigure}{0.45\textwidth}
    \centering
    \includegraphics[width=\linewidth]{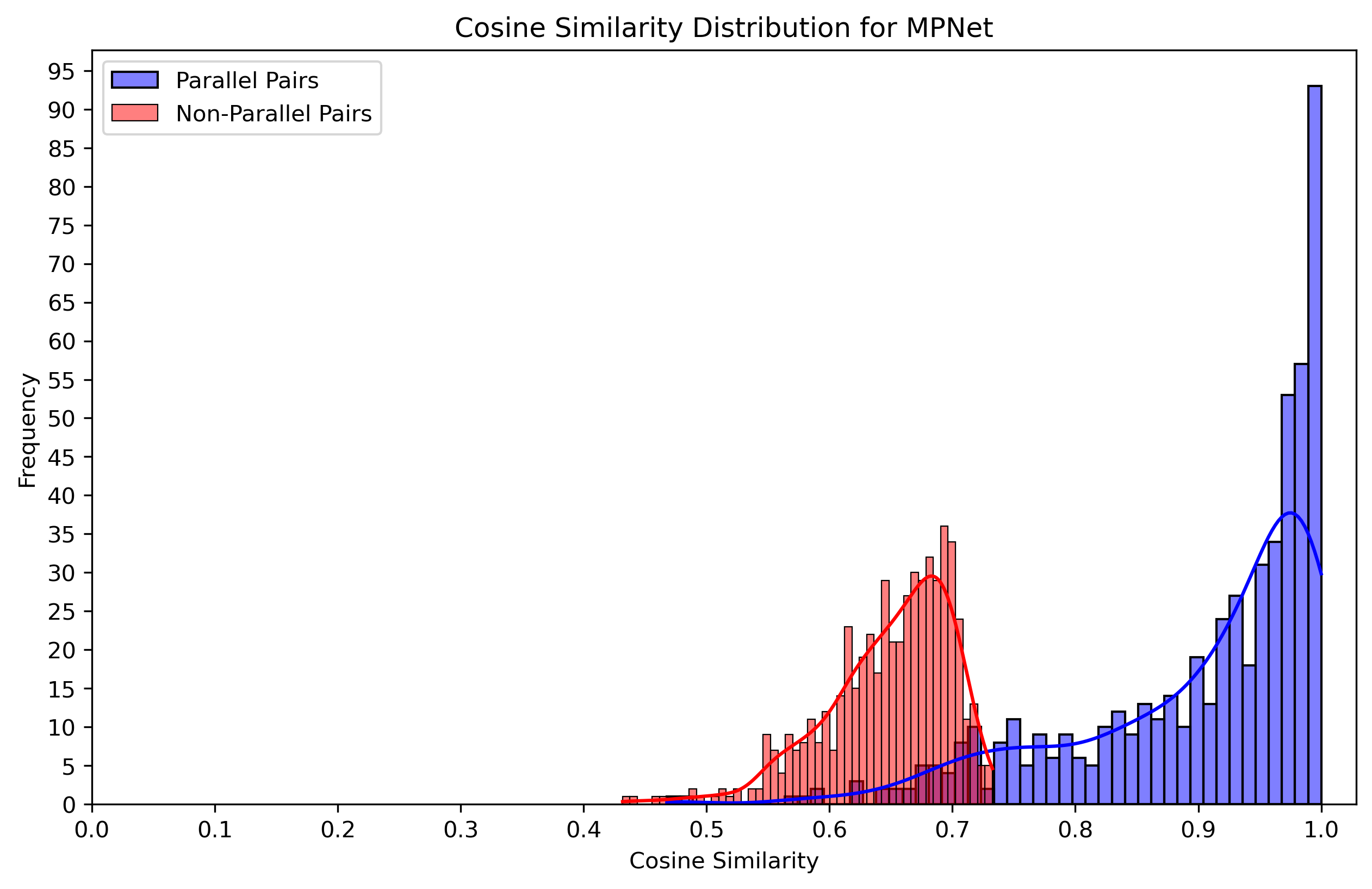}
    \caption{MPNet}
    \label{fig:mpnet_dist}
\end{subfigure}

\vspace{1em}
\caption{Distribution of cosine similarity scores for parallel (blue) and non-parallel (red) passages across different models.}
\label{fig:distributions}
\end{figure}

\end{document}